\documentclass[10pt,twocolumn,letterpaper]{article}

\usepackage[pagenumbers]{cvpr} %

\usepackage{graphicx}
\usepackage{amsmath}
\usepackage{amssymb}
\usepackage{booktabs}
\usepackage{bm}

\usepackage{subcaption}
\usepackage{booktabs}
\usepackage{multirow}
\usepackage{tabularx}
\usepackage{makecell}
\usepackage[usenames,dvipsnames,svgnames,table]{xcolor}
\usepackage{soul}
\usepackage{float}

\usepackage[pagebackref,breaklinks,colorlinks]{hyperref}

\usepackage[capitalize]{cleveref}
\crefname{section}{Sec.}{Secs.}
\Crefname{section}{Section}{Sections}
\Crefname{table}{Table}{Tables}
\crefname{table}{Tab.}{Tabs.}

\begin{document}

\title{GMFlow: Learning Optical Flow via Global Matching}

\author{{Haofei Xu\textsuperscript{1}\thanks{Work done while Haofei was at JD Explore Academy} \quad Jing Zhang\textsuperscript{2} \quad Jianfei Cai\textsuperscript{1}  \quad Hamid Rezatofighi\textsuperscript{1}\quad Dacheng Tao\textsuperscript{3,2}} \\
{\normalsize \textsuperscript{1}Department of Data Science and AI, Monash University, Australia}
\\
{\normalsize \textsuperscript{2}The University of Sydney, Australia}
\quad 
{\normalsize \textsuperscript{3}JD Explore Academy, China}\\
{\tt\footnotesize \{xuhfgm,dacheng.tao\}@gmail.com \ jing.zhang1@sydney.edu.au \ \{jianfei.cai,hamid.rezatofighi\}@monash.edu}
}

\maketitle

\begin{abstract}

Learning-based optical flow estimation has been dominated with the pipeline of cost volume with convolutions for flow regression, which is inherently limited to local correlations and thus is hard to address the long-standing challenge of large displacements. To alleviate this, the state-of-the-art framework RAFT gradually improves its prediction quality by using a large number of iterative refinements, achieving remarkable performance but introducing linearly increasing inference time. To enable both high accuracy and efficiency, we completely revamp the dominant flow regression pipeline by reformulating optical flow as a \emph{global matching} problem, which identifies the correspondences by directly comparing feature similarities. Specifically, we propose a GMFlow framework, which consists of three main components: a customized Transformer for feature enhancement, a correlation and softmax layer for global feature matching, and a self-attention layer for flow propagation. We further introduce a refinement step that reuses GMFlow at higher feature resolution for residual flow prediction. Our new framework outperforms 31-refinements RAFT on the challenging Sintel benchmark, while using only one refinement and running faster, suggesting a new paradigm for accurate and efficient optical flow estimation. Code is available at \href{https://github.com/haofeixu/gmflow}{https://github.com/haofeixu/gmflow}.

\end{abstract}

\vspace{-6pt}
\section{Introduction}
\label{sec:intro}
Since the pioneering learning-based work, FlowNet \cite{dosovitskiy2015flownet}, optical flow has been regressed with convolutions for a long time \cite{ilg2017flownet,ranjan2017optical,sun2018pwc,hur2019iterative,teed2020raft,xu2021high,Jiang_2021_ICCV,Zhang_2021_ICCV}. To encode the matching information into the network, the cost volume (\ie, correlation) \cite{hosni2012fast} was shown to be an effective component and thus has been extensively used in popular frameworks \cite{ilg2017flownet,sun2018pwc,teed2020raft}. However, such regression-based approaches have one major intrinsic limitation. That is, the cost volume requires a predefined size, as the search space is viewed as the channel dimension for subsequent regression with convolutions. This requirement restricts the search space to a \emph{local} range, making it hard to handle large displacements.

\begin{figure}[t]
    \centering
    \includegraphics[width=0.95\linewidth]{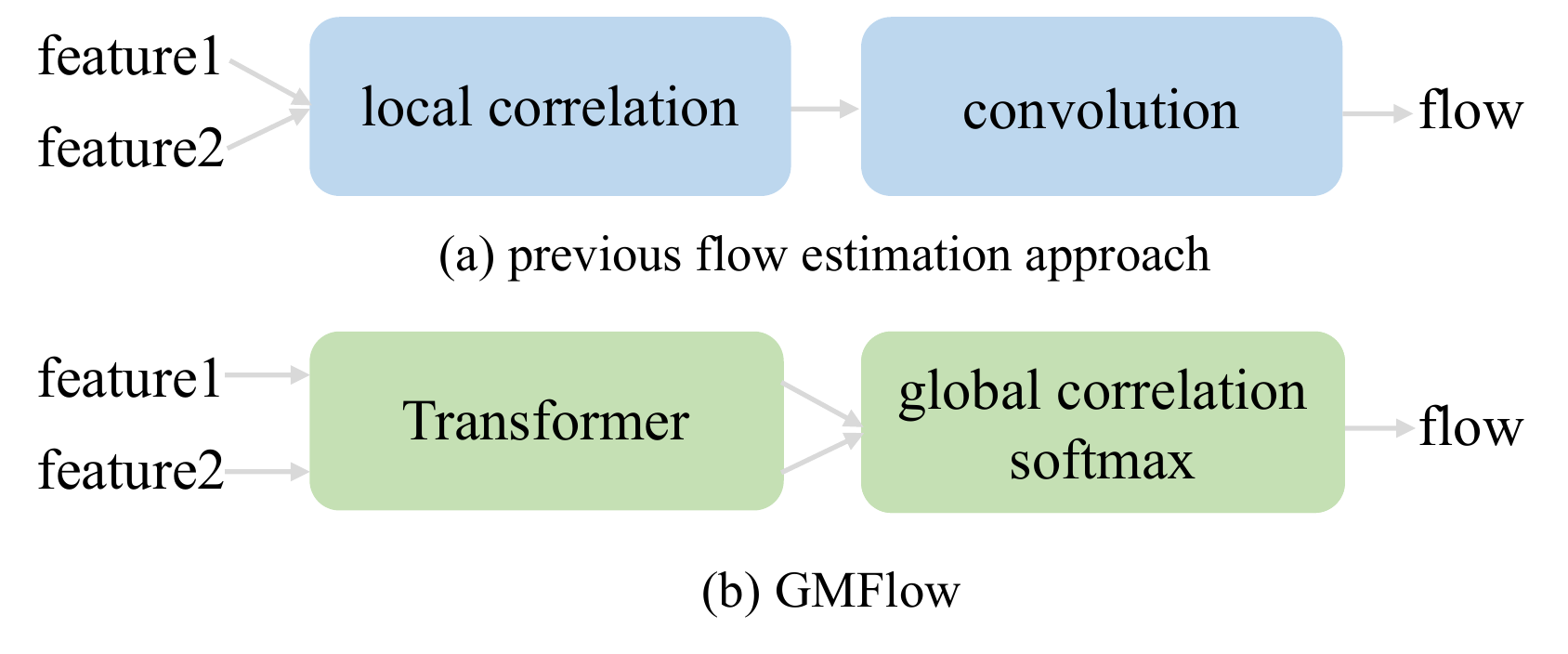}
    \vspace{-12pt}
    \caption{\textbf{Conceptual comparison of flow estimation approaches.} Most previous methods \emph{regress} optical flow from a \emph{local} cost volume (\ie, correlation) with convolutions, while we perform \emph{global matching} with a Transformer and differentiable matching layer (\ie, correlation and softmax).}
    \label{fig:flow_reg}
    \vspace{-12pt}
\end{figure}

To alleviate the large displacements issue, RAFT \cite{teed2020raft} proposes
an iterative framework with a large number of iterative refinements where convolutions are applied to different local cost volumes at different iteration stages so as to gradually reach \textit{near-global} search space, achieving outstanding performance on standard benchmarks. The current state-of-the-art methods \cite{Jiang_2021_ICCV, xu2021high, Zhang_2021_ICCV, teed2020raft} are all based on such an iterative architecture. Despite the excellent performance, such a large number of sequential refinements introduce linearly increasing inference time, which makes it hard for speed optimization and hinders its deployment in real-world applications. 
This raises a question: \textit{Is it possible to achieve both high accuracy and efficiency without requiring such a large number of iterative refinements?}

We notice that %
another visual correspondence problem, \ie, sparse matching between an image pair \cite{sarlin2020superglue,wang2020learning,sun2021loftr}, which usually features a \emph{large viewpoint change}, moves into a different track. %
The top-performing sparse methods (\eg, SuperGlue \cite{sarlin2020superglue} and LoFTR \cite{sun2021loftr}) adopt \emph{Transformers} \cite{vaswani2017attention} to reason about the mutual relationship between feature descriptors, and the correspondences are extracted with an explicit \emph{matching} layer (\eg, softmax layer \cite{wang2020learning}).

Inspired by sparse matching, we propose to completely remove the additional convolutional layers operating on a predefined local cost volume, and reformulate optical flow as a \emph{global matching} problem, which is distinct from all previous learning-based optical flow methods. Fig.~\ref{fig:flow_reg} provides a conceptual comparison of these two flow estimation approaches. Our flow prediction is obtained with a differentiable matching layer, \ie, correlation and softmax layer, by comparing feature similarities. Such a formulation calls for more discriminative feature representations, for which the Transformer \cite{vaswani2017attention} becomes a natural choice.

We would like to point out that although our pipeline shares the conceptual major components (\ie, Transfromer and the softmax matching layer) with sparse matching~\cite{sarlin2020superglue,sun2021loftr}, our motivation is originated from the development of optical flow methods and the challenges associated with formulating optical flow as a global matching problem are quite different. Optical flow focuses on dense correspondences for every pixel and modern learning-based architectures are mostly designed by regression from a local cost volume. The scale and complexity of optical flow are much higher. Moreover, optical flow needs to deal with occluded and out-of-boundary pixels, for which the simple softmax-based matching layer will not be effective.

In this paper, we propose a GMFlow framework to realize the global matching formulation for optical flow. Specifically, the dense features extracted from a convolutional backbone network are fed into a Transformer that consists of self-, cross-attentions and feed-forward network to obtain more discriminative features. We then compare the feature similarities by correlating all pair-wise features.
After that, the flow prediction is obtained with a differentiable softmax matching layer. 
To address occluded and out-of-boundary pixels, %
we incorporate an additional self-attention layer to propagate the high-quality flow prediction from matched %
pixels to unmatched ones %
by exploiting the feature self-similarity \cite{hui2020liteflownet3,Jiang_2021_ICCV}. %
We further introduce a refinement step that reuses GMFlow at higher feature resolution for residual flow prediction. Our full framework achieves competitive performance and higher efficiency compared with the state-of-the-art methods. Specifically, with only one refinement, GMFlow outperforms 31-refinements RAFT on the challenging Sintel \cite{butler2012naturalistic} dataset, while running faster.

Our major contributions can be summarized as follows:
\begin{itemize}
	\vspace{-8pt}
	\item We completely revamp the dominant flow regression pipeline by reformulating optical flow as a global matching problem, which effectively addresses the long-standing challenge of large displacements.
	\vspace{-8pt}
	\item We propose a GMFlow framework to realize the global matching formulation, which consists of three main components: a Transformer for feature enhancement, a correlation and softmax layer for global feature matching, and a self-attention layer for flow propagation.
	\vspace{-8pt}
	\item We further propose a refinement step to exploit higher resolution feature, which allows us to reuse the same GMFlow framework for residual flow estimation.
	\vspace{-8pt}
	\item GMFlow outperforms 31-refinements RAFT on the challenging Sintel benchmark, while using only one refinement and running faster, suggesting a new paradigm for accurate and efficient flow estimation.
	\vspace{-8pt}
\end{itemize}

\begin{figure*}
    \centering
    \includegraphics[width=0.95\linewidth]{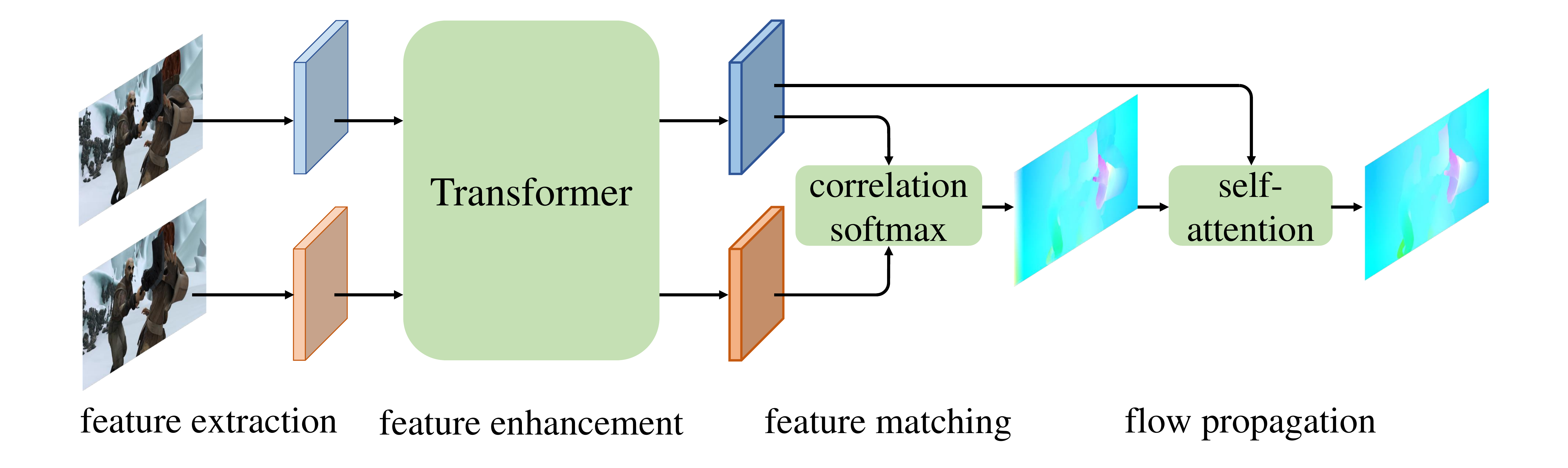}
    \vspace{-14pt}
    \caption{\textbf{Overview of GMFlow framework}. We first extract $8\times$ downsampled dense features from two input video frames with a weight-sharing convolutional network. Then the features are fed into a Transformer for feature enhancement. Next we compare feature similarities by correlating all pair-wise features and the optical flow is obtained with a softmax matching layer. An additional self-attention layer is introduced to propagate the high-quality flow predictions in matched pixels to unmatched ones by considering the feature self-similarity.}
    \label{fig:overview}
    \vspace{-14pt}
\end{figure*}

\section{Related Work}

{\bf Flow estimation approach.} The flow estimation approach is fundamental to existing popular optical flow frameworks \cite{ilg2017flownet,sun2018pwc,hur2019iterative,teed2020raft,xu2021high,Jiang_2021_ICCV,Zhang_2021_ICCV}, notably the coarse-to-fine method PWC-Net \cite{sun2018pwc} and iterative refinement method RAFT \cite{teed2020raft}. They both perform some sort of multi-stage refinements, either at multiple scales \cite{sun2018pwc} or a single resolution \cite{teed2020raft}. For flow prediction at each stage, their pipeline is conceptual similar, \ie, regressing optical flow from a local cost volume with convolutions. However, such an approach is hard to handle large displacements. Thus multi-stage refinements are required to estimate large motion incrementally. The success of RAFT largely lies in the large number of iterative refinements it can perform. Two exceptions to these local regression approaches are DICL \cite{wang2020displacement} and GLU-Net \cite{truong2020glu}. DICL performs \emph{local matching} with \emph{convolutions}, while GLU-Net \emph{regresses} flow from \emph{global correlation} with \emph{convolutions}, which also makes itself restricted to fixed image resolution and thus an additional sub-network is required to handle this issue. Distinct from these approaches, we perform \emph{global matching} with a \emph{Transformer} and we show it is indeed possible to achieve highly accurate results without relying on a large number of refinements.

{\bf Large displacements.} Large displacements have been a long-standing challenge for optical flow \cite{brox2009large,weinzaepfel2013deepflow,brox2004high,chen2013large,bailer2015flow}.
One popular strategy is to use the coarse-to-fine approach \cite{hu2016efficient,brox2004high,sun2018pwc} that estimates large motion incrementally. However, coarse-to-fine methods tend to miss fast-moving small objects \cite{revaud2015epicflow} if the resolution is too coarse. To alleviate this issue, RAFT \cite{teed2020raft} proposes to maintain a single high resolution and gradually improve the initial prediction with a large number of iterative refinements, achieving remarkable performance on standard benchmarks. However, such a large number of sequential refinements introduce linearly increasing inference time, which makes it hard for speed optimization and hinders its integration into real-world systems. In contrast, our new framework streamlines the optical flow pipeline and estimates large displacements with both high accuracy and efficiency, achieved by a reformulation of the optical flow problem and a strong Transformer.

{\bf Transformer for correspondences.} SuperGlue \cite{sarlin2020superglue} pioneered the use of Transformer for sparse feature matching. LoFTR \cite{sun2021loftr} further improves its performance by removing the feature detection step in the typical pipelines. COTR \cite{jiang2021cotr} formulates the correspondence problem as a functional mapping by querying the interest point and uses a Transformer as the function. These frameworks are mainly designed for sparse matching problems, and it is non-trivial to directly adopt them to dense correspondence tasks. Although COTR in principle can also predict dense correspondences by querying every pixel location, the inference speed will be significantly slower. For dense correspondences, STTR \cite{li2021revisiting} is a Transformer-based method for stereo matching, which can be viewed as a special case of optical flow. Besides, STTR relies on a complex optimal transport matching layer and doesn't produce predictions for occluded pixels, while we use a much simpler softmax operation and a simple flow propagation layer to handle occlusion. Another related work Perceiver IO \cite{jaegle2021perceiver} targets general problems and uses optical flow as an application in experiments. Although both method are based on Transformers, Perceiver IO directly regresses optical flow via self-attention, while we aim at learning strong feature representations (in particular with cross-attention) for matching.

\section{Methodology}
Optical flow is intuitively a matching problem that aims at finding corresponding pixels. To achieve this, one can compare the similarity of the features for each pixel, and identify the corresponding pixel that gives the highest similarity. Such a process requires the features to be discriminative enough to stand out. Aggregating the spatial contexts within the image itself and information from another image can intuitively alleviate ambiguities and improve their distinctiveness. Such design philosophies have enabled great achievement in sparse feature matching frameworks \cite{sarlin2020superglue,sun2021loftr}. The success of sparse matching, which usually features a large viewpoint change, motivates us to formulate optical flow as an explicit global matching problem in order to address the challenge of large displacements.

In the following, we first provide a general description of our global matching formulation, and then present a Transformer-based framework to realize it.

\subsection{Formulation}
\label{sec:formulation}
Given two consecutive video frames ${\bm I}_1$ and ${\bm I}_2$, we first extract downsampled dense features ${\bm F}_1, {\bm F}_2 \in \mathbb{R}^{H \times W \times D}$ with a weight-sharing convolutional network, where $H, W$ and $D$ denote height, width and feature dimension, respectively. Considering the correspondences in the two frames should share high similarity, we first compare the feature similarity for each pixel in ${\bm F}_1$ with respect to all pixels in ${\bm F}_2$ by computing their correlations \cite{wang2020learning}. This can be implemented efficiently with a simple matrix multiplication:
\begin{equation}
\label{eq:corr}
    {\bm C} =  \frac{{\bm F}_1 {\bm F}_2^T}{\sqrt{D}}  \in \mathbb{R}^{H \times W \times H \times W}, 
\end{equation}
where each element in the correlation matrix ${\bm C}$ represents the correlation value between coordinates ${\bm p}_1 = (i, j)$ in ${\bm F}_1$ and ${\bm p}_2 = (k, l)$ in ${\bm F}_2$, and $\frac{1}{\sqrt{D}}$ is a normalization factor to avoid large values after the dot-product operation \cite{vaswani2017attention}.

To identify the correspondence, one na\"ive approach is to directly take the location that gives the highest correlation. However, this operation is unfortunately non-differentiable, which prevents end-to-end training. To tackle this issue, we use a differentiable matching layer \cite{wang2020learning,kendall2017end,xu2020aanet}. Specifically, we normalize the last two dimensions of ${\bm C}$ with the softmax operation, which gives us a matching distribution
\begin{equation}
    {\bm M} = \mathrm{softmax} ({\bm C}) \in \mathbb{R}^{H \times W \times H \times W}
\end{equation}
for each location in ${\bm F}_1$ with respect to all locations in ${\bm F}_2$. Then, the correspondence $\hat{{\bm G}}$ can be obtained by taking a weighted average of the 2D coordinates of pixel grid ${\bm G} \in \mathbb{R}^{H \times W \times 2}$ with the matching distribution ${\bm M}$:
\begin{equation}
    \hat{{\bm G}} = {\bm M} {\bm G} \in \mathbb{R}^{H \times W \times 2}.
\end{equation}
Finally, the optical flow ${\bm V}$ can be obtained by computing the difference between the corresponding pixel coordinates
\begin{equation}
{\bm V} = \hat{{\bm G}}  - {\bm G} \in \mathbb{R}^{H \times W \times 2}.
\end{equation}
Such a softmax-based approach can not only enable end-to-end training but also provide sub-pixel accuracy.

\subsection{Feature Enhancement}

Key to our formulation lies in obtaining high-quality discriminative features for matching. Recall that the features ${\bm F}_1$ and ${\bm F}_2$ in Sec.~\ref{sec:formulation} are extracted \emph{independently} from a weight-sharing convolutional network. To further consider their mutual dependencies, a natural choice is Transformer \cite{vaswani2017attention}, which is particularly suitable for modeling the mutual relationship between two sets with the attention mechanism, as demonstrated in sparse matching methods \cite{sarlin2020superglue,sun2021loftr}.
Since ${\bm F}_1$ and ${\bm F}_2$ are only two sets of features, they have no notion of the spatial position, we first add the fixed 2D sine and cosine positional encodings (following DETR \cite{carion2020end}) to the features. Adding the position information also makes the matching process consider not only the feature similarity but also their spatial distance, which can help resolve ambiguities and improve the performance (Table~\ref{tab:transformer}).

After adding the position information, we perform six stacked self-, cross-attentions and feed-forward network (FFN) \cite{vaswani2017attention} to improve the quality of the initial features. Specifically, for self-attention, the query, key and value in the attention mechanism \cite{vaswani2017attention} are the same feature. For cross-attention, the key and value are same but different from the query to introduce their mutual dependencies. This process is performed for both ${\bm F}_1$ and ${\bm F}_2$ symmetrically, \ie,
\begin{equation}
\label{eq:trans_feature}
    \hat{\bm F}_1 = \mathcal{T}({\bm F}_1 + {\bm P}, {\bm F}_2 + {\bm P}), \hspace{0.6em} \hat{\bm F}_2 = \mathcal{T}({\bm F}_2 + {\bm P}, {\bm F}_1 + {\bm P}),
\end{equation}
where $\mathcal{T}$ is a Transformer, ${\bm P}$ is the positional encoding, the first input of $\mathcal{T}$ is query and the second is key and value.

One issue in the standard Transformer architecture \cite{vaswani2017attention} is the quadratic computational complexity due to the pair-wise attention operation. To improve the efficiency, we adopt the shifted local window attention strategy from Swin Transformer \cite{liu2021Swin}. However, unlike Swin that uses \emph{fixed window size}, we split the feature to \emph{fixed number of local windows} to make the window size adaptive with the feature size. Specifically, we split the input feature of size $H \times W$ to $K \times K$ windows (each with size $\frac{H}{K} \times \frac{W}{K}$), and perform self- and cross-attentions within each local window independently. For every two consecutive local windows, we shift the window partition by $(\frac{H}{2K}, \frac{W}{2K})$ to introduce cross-window connections. In our framework, we split to $2 \times 2$ windows (each with size $\frac{H}{2} \times \frac{W}{2}$), which represents a good speed-accuracy trade-off (Table~\ref{tab:split_attn}).

\begin{figure}[t]
    \centering
    \includegraphics[width=0.98\linewidth]{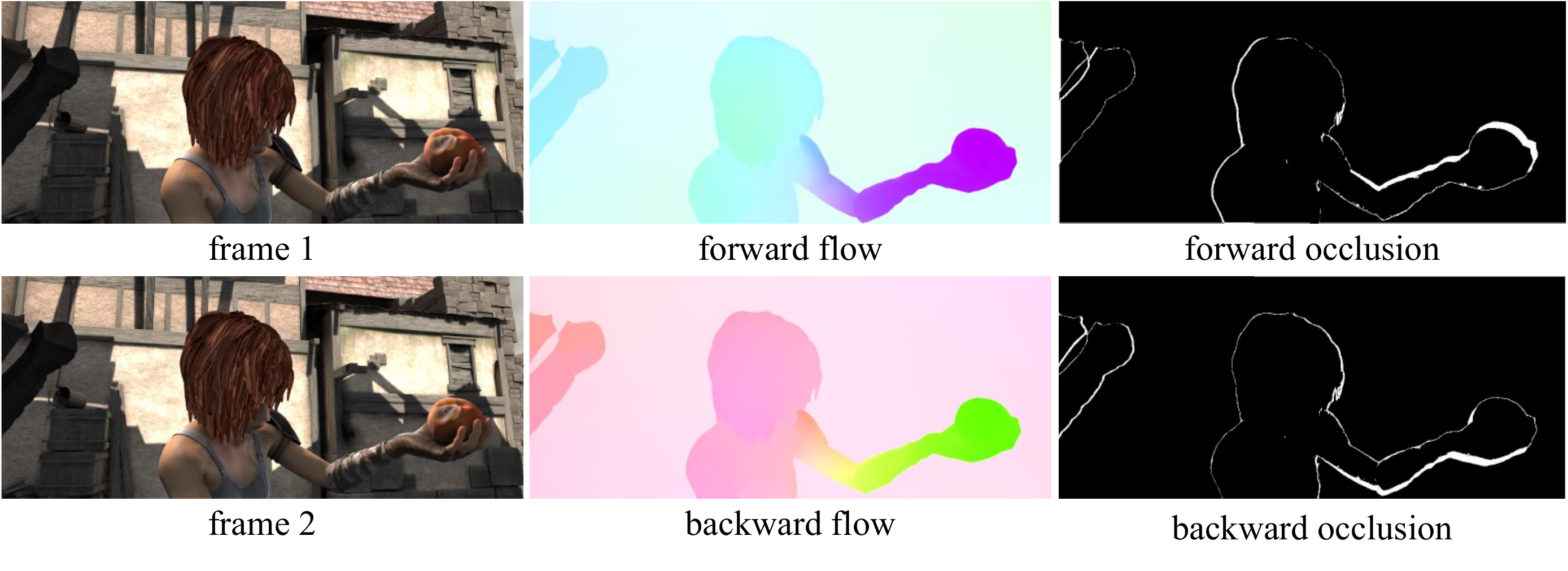}
    \vspace{-6pt}
    \caption{\textbf{GMFlow also simplifies backward flow computation} by transposing the global correlation matrix without requiring to forward the network twice. The bidirectional flow can be used for occlusion detection with forward-backward consistency check.}
    \label{fig:bidir_flow}
    \vspace{-12pt}
\end{figure}

\subsection{Flow Propagation}
Our softmax-based flow estimation method implicitly assumes that the corresponding pixels are visible in both images and thus they can be matched by comparing their similarities. However, this assumption will be invalid for occluded and out-of-boundary pixels. To remedy this, by observing that the optical flow field and the image itself share high structure similarity \cite{hui2020liteflownet3,Jiang_2021_ICCV}, we propose to propagate the high-quality flow predictions in matched pixels to unmatched ones by measuring the feature self-similarity. This operation can be implemented efficiently with a simple self-attention layer (illustrated in Fig.~\ref{fig:overview}):
\begin{equation}
    \Tilde{\bm V} = \mathrm{softmax} \left(\frac{\hat{\bm F}_1 \hat{\bm F_1}^T}{\sqrt{D}}\right) \hat{\bm V}  \in \mathbb{R}^{H \times W \times 2},
\end{equation}
where
\begin{equation}
    \hat{\bm V} = \mathrm{softmax} \left(\frac{\hat{\bm F}_1 \hat{\bm F_2}^T}{\sqrt{D}}\right) {\bm G} - {\bm G},
\end{equation}
is the optical flow prediction from the softmax layer, which is obtained by substituting the stronger features in Eq.~\eqref{eq:trans_feature} into the softmax matching layer in Sec.~\ref{sec:formulation}. Fig.~\ref{fig:overview} provides an overview of our GMFlow framework.

\begin{table*}[t]
    \centering
    \setlength{\tabcolsep}{2.2pt} %
    \begin{tabular}{lcccccccccccccccccc}
    \toprule
    \multirow{2}{*}[-2pt]{Method} & \multirow{2}{*}[-2pt]{\#blocks} & \multicolumn{4}{c}{Things (val, clean)} & \multicolumn{4}{c}{Sintel (train, clean)} & \multicolumn{4}{c}{Sintel (train, final)} &  \multirow{2}{*}[-2pt]{\begin{tabular}[x]{@{}c@{}}Param\\(M) \end{tabular}} \\
    \addlinespace[-12pt] \\
    \cmidrule(lr){3-6} \cmidrule(lr){7-10} \cmidrule(lr){11-14}
    \addlinespace[-12pt] \\
    & & EPE & $s_{0-10}$ & $s_{10-40}$ & $s_{40+}$ & EPE & $s_{0-10}$ & $s_{10-40}$ & $s_{40+}$ & EPE & $s_{0-10}$ & $s_{10-40}$ & $s_{40+}$ & \\
    
    \midrule

    \multirow{4}{*}[-2pt]{cost volume + conv} & 0 & 18.83 & 3.42 & 6.49 & 49.65 & 6.45 & 1.75 & 7.17 & 38.19 & 7.75 & 2.10 & 8.88 & 45.29 & 1.8 \\
    & 4 & 10.99 & 1.70 & 3.41 & 29.78 & 3.32 & 0.73 & 3.84 & 20.58 & 4.93 & 0.99 & 5.71 & 31.16 & 4.6 \\
    & 8 & 9.59 & 1.44 & 2.96 & 26.04 & 2.89 & 0.65 & 3.36 & 17.75 & 4.32 & 0.88 & 4.95 & 27.33 & 8.0 \\
    & 12 & 9.04 & 1.37 & 2.84 & 24.46 & 2.78 & 0.65 & 3.32 & 16.69 & 4.07 & 0.84 & 4.76 & 25.44 & 11.5 \\
    & 18 & 8.67 & 1.33 & 2.74 & 23.43 & 2.61 & 0.59 & 3.07 & 15.91 & 3.94 & 0.82 & 4.62 & 24.58 & 15.7 \\

    \midrule
    
    \multirow{4}{*}[-2pt]{Transformer + softmax} & 0 & 22.93 & 8.57 & 11.13 & 52.07 & 8.44 & 2.71 & 11.60 & 42.10 & 10.28 & 3.11 & 13.83 & 53.34 & 1.0 \\
    & 1 & 11.45 & 2.98 & 4.68 & 28.35 & 4.12 & 1.27 & 5.08 & 22.25 & 6.11 & 1.70 & 7.89 & 33.52 & 1.6 \\
    & 2 & 8.59 & 1.80 & 3.28 & 21.99 & 3.09 & 0.90 & 3.66 & 17.37 & 4.54 & 1.24 & 5.44 & 26.00 &  2.1 \\  %
    & 4 & 7.19 & 1.40 & 2.62 & 18.66 & 2.43 & 0.67 & 2.73 & 14.23 & 3.78 & 1.01 & 4.27 & 22.37 & 3.1 \\  %
    & 6 & \textbf{6.67} & \textbf{1.26} & \textbf{2.40} & \textbf{17.37} & \textbf{2.28} & \textbf{0.58} & \textbf{2.49} & \textbf{13.89} & \textbf{3.44} & \textbf{0.80} & \textbf{3.97} & \textbf{21.02} & 4.2 \\  %
    
    \bottomrule
    \end{tabular}
    \vspace{-5pt}
    \caption{\textbf{Methodology comparison}. We stack different number of convolutional residual blocks or Transformer blocks to see how the performance varies. All models are trained on Chairs and Things training sets. We report the performance on Things validation clean set and the cross-dataset generalization results on Sintel training clean and final sets. 
    }
    \label{tab:conv_vs_softmax}
    \vspace{-6pt}
\end{table*}

\subsection{Refinement}
\label{sec:refine}

The framework presented so far (based on $1/8$ features) can already achieve competitive performance (Table~\ref{tab:raft_vs_ours}). It can be further improved by introducing additional higher resolution ($1/4$) feature for refinement. Specifically, we first upsample the previous $1/8$ flow prediction to $1/4$ resolution, and warp the second feature with the current flow prediction. Then the refinement task is reduced to the residual flow learning, where the same GMFlow framework depicted in Fig.~\ref{fig:overview} can be used but in a local range. Specifically, we split to $8 \times 8$ local windows (each with $1/32$ of the original image resolution) in the Transformer and perform a $9 \times 9$ local window matching for each pixel. After obtaining the flow prediction from the softmax layer, we perform a $3 \times 3$ local window self-attention operation for flow propagation.

Note that here we share the Transformer and self-attention weights in the refinement step with the global matching stage, which not only reduces parameters but also improves the generalization (Table~\ref{tab:refine}). To generate the $1/4$ and $1/8$ features, we also share the backbone features. Specifically, we take a similar approach to TridentNet \cite{li2019scale} but use a weight-sharing convolution with strides $1$ and $2$, respectively. Such a weight-sharing design also leads to better performance than the feature pyramid network \cite{lin2017feature}.

\subsection{Training Loss}
We supervise all flow predictions using $\ell_1$ loss between the ground truth:
\begin{equation}
    L = \sum_{i=1}^{N} \gamma^{N - i} \| {\bm V}_i - {\bm V}_{\mathrm{gt}} \|_1,
\end{equation}
where $N$ is the number of flow predictions including the intermediate and final ones, and $\gamma$ (set to 0.9) is the weight that is exponentially increasing to give higher weights for later predictions following RAFT \cite{teed2020raft}.

\begin{table*}[t]
\centering
\subfloat[
\textbf{Transformer components}. Cross-attention contributes most.
\label{tab:transformer}
]{
\centering
\begin{minipage}{0.45\linewidth}
{\begin{center}
\begin{tabular}{lccccccccccccccc}
    \toprule
    
    \multirow{2}{*}[-2pt]{setup} & \multicolumn{1}{c}{Things (val)} & \multicolumn{2}{c}{Sintel (train)} &  \multirow{2}{*}[-2pt]{\begin{tabular}[x]{@{}c@{}}Param\\(M) \end{tabular}} \\
    \addlinespace[-12pt] \\
    \cmidrule(lr){2-2} \cmidrule(lr){3-4}
    \addlinespace[-12pt] \\
    & clean & clean & final & \\
    \midrule

    full & \textbf{6.67} & \textbf{2.28} & \textbf{3.44} & 4.2 \\
    
    w/o cross & 10.84 & 4.48 & 6.32 & 3.8  \\
    w/o pos & 8.38 & 2.85 & 4.28 & 4.2 \\
    w/o FFN & 8.71 & 3.10 & 4.43 & 1.8 & \\
    w/o self & 7.04 & 2.49 & 3.69 & 3.8 \\
    
    \bottomrule
    \end{tabular}
\end{center}}
\end{minipage}
}
\hspace{2em}
\subfloat[
\textbf{Numbers of window splits in shifted local attention}. $2 \times 2$ represents a good speed-accuracy trade-off.
\label{tab:split_attn}
]{
\begin{minipage}{0.45\linewidth}{\begin{center}
\begin{tabular}{cccccccccccccccc}
    \toprule
    
    \multirow{2}{*}[-2pt]{\#splits} & \multicolumn{4}{c}{Things (val, clean)}  &  \multirow{2}{*}[-2pt]{\begin{tabular}[x]{@{}c@{}}Time\\(ms) \end{tabular}} \\
    \addlinespace[-12pt] \\
    \cmidrule(lr){2-5} 
    \addlinespace[-12pt] \\
    & EPE & $s_{0-10}$ & $s_{10-40}$ & $s_{40+}$  \\
    \midrule
    
    $1 \times 1$ & 6.34 & 1.26 & 2.37 & 16.36 & 105  \\
    \underline{$2 \times 2$} & 6.67 & 1.26 & 2.40 & 17.37 & 53 \\
    $4 \times 4$ & 7.32 & 1.29 & 2.58 & 19.26 & 35 \\
    \bottomrule
    \\
    \\

    \end{tabular}
\end{center}}\end{minipage}
}
\\
\subfloat[
\textbf{Global \vs local matching}. Global matching is significantly better for large motion and is fast to compute.
\label{tab:global_local_match}
]{
\begin{minipage}{0.45\linewidth}{\begin{center}
\setlength{\tabcolsep}{3pt} %
\begin{tabular}{ccccccccccccccc}
    \toprule
    
    \multirow{2}{*}[-2pt]{\begin{tabular}[x]{@{}c@{}}matching \\space \end{tabular}} & \multicolumn{4}{c}{Things (val, clean)}  & \multirow{2}{*}[-2pt]{\begin{tabular}[x]{@{}c@{}}Time\\(ms) \end{tabular}} \\
    \addlinespace[-12pt] \\
    \cmidrule(lr){2-5} 
    \addlinespace[-12pt] \\
    & EPE & $s_{0-10}$ & $s_{10-40}$ & $s_{40+}$  \\
    \midrule
    
    global & \textbf{6.67} & 1.26 & \textbf{2.40} & \textbf{17.37} & 52.6 \\
    local $3 \times 3$ & 31.78 & 1.19 & 12.40 & 85.39 & 51.2 \\
    local $5 \times 5$ & 26.51 & \textbf{0.89} & 6.67 & 76.76 & 51.5 \\
    local $ 9 \times 9$ & 19.88 & 1.01 & 2.44 & 61.06 & 52.9 \\

    \bottomrule
    \end{tabular}
\end{center}}\end{minipage}
}
\hspace{2em}
\subfloat[
\textbf{Flow propagation} greatly improves unmatched pixels. 
\label{tab:prop}
]{
\begin{minipage}{0.45\linewidth}{\begin{center}
\setlength{\tabcolsep}{1pt} %
\begin{tabular}{lccccccccccccccc}
    \toprule
    
    \multirow{2}{*}[-2pt]{prop.} & \multicolumn{3}{c}{Sintel (clean)}  &  \multicolumn{3}{c}{Sintel (final)} \\
    \addlinespace[-12pt] \\
    \cmidrule(lr){2-4} \cmidrule(lr){5-7} 
    \addlinespace[-12pt] \\
    & all & matched & unmatched & all & matched & unmatched  \\
    \midrule
    
    w/o  & 2.28 & \textbf{1.06} & 15.54 & 3.44 & \textbf{1.95} & 19.50  \\
    w/  & \textbf{1.89} & 1.10 & \textbf{10.39} & \textbf{3.13} & 1.98 & \textbf{15.52} \\

    \bottomrule
    
    \\
    
    \\
    \end{tabular}
\end{center}}\end{minipage}
}
\\
\subfloat[
\textbf{Sharing Transformer and multi-scale features} leads to better performance and less parameters.
\label{tab:refine}
]{
\centering
\begin{minipage}{0.45\linewidth}{\begin{center}
\setlength{\tabcolsep}{1.5pt} %
    \begin{tabular}{lccccccccccccccc}
    \toprule
    
    \multirow{2}{*}[-2pt]{setup} & \multirow{2}{*}[-2pt]{\begin{tabular}[x]{@{}c@{}}share\\Trans? \end{tabular}}    & \multirow{2}{*}[-2pt]{\begin{tabular}[x]{@{}c@{}}share\\feature? \end{tabular}} & \multicolumn{1}{c}{Things} & \multicolumn{2}{c}{Sintel (train)} &  \multirow{2}{*}[-2pt]{\begin{tabular}[x]{@{}c@{}}Param\\(M) \end{tabular}}  \\
    \addlinespace[-12pt] \\
    \cmidrule(lr){4-4} \cmidrule(lr){5-6}
    \addlinespace[-12pt] \\
    &  &  & clean & clean & final &  & \\

    \midrule
    
    w/o refine & - & - & 3.98 & 1.65 & 2.94 & 4.7 \\
    
    \midrule
    
    \multirow{4}{*}[-2pt]{w/ refine} & & & 3.41 & 1.28 & 2.75 & 8.0 \\
    & \checkmark & & \textbf{3.21} & 1.27 & 2.60 & 4.9 \\
    & & \checkmark & 3.26 & 1.21 & 2.70 & 7.9 \\
    & \checkmark & \checkmark & 3.24 & \textbf{1.20} & \textbf{2.59} & \textbf{4.7} \\

    \bottomrule
    \end{tabular}
\end{center}}\end{minipage}
}
\hspace{2em}
\subfloat[
\textbf{Training length}. GMFlow benefits from more training iterations.
\label{tab:train_schedule}
]{
\begin{minipage}{0.45\linewidth}{\begin{center}
\setlength{\tabcolsep}{3.pt} %
    \begin{tabular}{cccccccccccccccc}
    \toprule
    
    \multirow{2}{*}[-2pt]{\begin{tabular}[x]{@{}c@{}}training \\iterations \end{tabular}} & \multicolumn{1}{c}{Things (val)} & \multicolumn{2}{c}{Sintel (train)} &  \multicolumn{2}{c}{KITTI (train)} \\
    \addlinespace[-12pt] \\
    \cmidrule(lr){2-2} \cmidrule(lr){3-4} \cmidrule(lr){5-6}
    \addlinespace[-12pt] \\
    & clean & clean & final & EPE & F1-all \\
    
    \midrule
    
    200K & 3.24 & 1.20 & 2.59 & 9.32 & 26.95 \\
    400K & 3.01 & 1.16 & 2.51 & 8.96 & 26.18 \\
    600K & 2.93 & 1.13 & \textbf{2.42} & 8.34 & 24.77 \\
    800K & \textbf{2.80} & \textbf{1.08} & 2.48 & \textbf{7.77} & \textbf{23.40} \\
    
    \bottomrule
    
    \\
    \\[-7pt]
    
    \end{tabular}
\end{center}}\end{minipage}
}
\vspace{-10pt}
\caption{\textbf{GMFlow ablations}. All models are trained on Chairs and Things training sets. }
\label{tab:ablations}
\vspace{-10pt}
\end{table*}

\section{Experiments}

{\bf Datasets and evaluation setup.} Following previous methods \cite{ilg2017flownet,sun2018pwc,teed2020raft}, we first train on the FlyingChairs (Chairs) \cite{dosovitskiy2015flownet} and FlyingThings3D (Things) \cite{mayer2016large} datasets, and then evaluate Sintel \cite{butler2012naturalistic} and KITTI \cite{menze2015object} training sets. We also evaluate on the Things validation set to see how the model performs on the same-domain data. Finally, we perform additional fine-tuning on Sintel and KITTI training sets and report the performance on the online benchmarks.

{\bf Metrics.} We adopt the commonly used metric in optical flow, \ie, the end-point-error (EPE), which is the average $\ell_2$ distance between the prediction and ground truth. For KITTI dataset, we also use \emph{F1-all}, which reflects the percentage of outliers. To better understand the performance gains, we also report the EPE in different motion magnitudes. 
Specifically, we use $s_{0-10}, s_{10-40}$ and $s_{40+}$, to denote the EPE over pixels with ground truth flow motion magnitude falling to $0-10$, $10-40$ and more than $40$ pixels.

{\bf Implementation details.} We implement our framework in PyTorch. Our convolutional backbone network is identical to RAFT's model, except that our final feature dimension is 128, while RAFT's is 256. We stack 6 Transformer blocks. To upsample the low-resolution flow prediction to the original image resolution, we use RAFT's convex upsampling \cite{teed2020raft} method. We use AdamW \cite{loshchilov2017decoupled} as the optimizer. We first train the model on Chairs dataset for 100K iterations, with a batch size of 16 and a learning rate of 4e-4. We then fine-tune it on Things dataset for 200K iterations for ablation experiments, with a batch size of 8 and a learning rate of 2e-4. Our best model is fine-tuned on Things for 800K iterations. For the final fine-tuning process on Sintel and KITTI datasets, we report the details in Sec.~\ref{sec:sintel_kitti}. Further details are provided in the \emph{supplementary material}.

\subsection{Methodology Comparison}
\label{sec:method_comp}

{\bf Flow estimation approach.} We compare our Transformer and softmax-based flow estimation method with the cost volume and convolution-based approach. Specifically, we adopt the state-of-the-art cost volume construction method in RAFT \cite{teed2020raft} that concatenates 4 local cost volumes at 4 scales, where each cost volume has a dimension of $H \times W \times (2R+1)^2$, where $H$ and $W$ are image height and width, respectively, and the search range $R$ is set to 4 following RAFT. To regress flow, we stack different number of convolutional residual blocks \cite{he2016deep} to see how the performance varies. The final optical flow is obtained with a $3 \times 3$ convolution with 2 output channels. For our proposed framework, we stack different number of Transformer blocks for feature enhancement, where one Transformer block consists of self-, cross-attentions and a feed-forward network (FFN). The final optical flow is obtained with a global correlation and softmax layer. Both methods use bilinear upsampling in this comparison. Table~\ref{tab:conv_vs_softmax} shows that the performance improvement of our method is more significant compared to the cost volume and convolution-based approach. For instance, our method with 2 Transformer blocks can already outperform 8 convolution blocks, especially for large motion ($s_{40+}$). The performance can be further improved by stacking more layers, surpassing the cost volume and convolution-based approach by a large margin. We present more comparisons with all the possible combinations of different flow estimation approaches in the \emph{supplementary material}, where our method is consistently better and has less parameters than other variants.

{\bf Bidirectional flow prediction.} Our framework also simplifies backward optical flow computation by directly transposing the global correlation matrix in Eq.~\eqref{eq:corr}. Note that during training we only predict unidirectional flow while at inference we can obtain bidirectional flow for free, without requiring to forward the network twice, unlike previous regression-based methods \cite{meister2018unflow,hur2019iterative}. The bidirectional flow can be used for occlusion detection with forward-backward consistency check (following \cite{meister2018unflow}), as shown in Fig.~\ref{fig:bidir_flow}.

\begin{table*}[t]
    \centering
    \setlength{\tabcolsep}{3.pt} %
    \begin{tabular}{lccccccccccccccr}
    \toprule
    \multirow{2}{*}[-2pt]{Method} & \multirow{2}{*}[-2pt]{\#refine.} & \multicolumn{4}{c}{Things (val, clean)} & \multicolumn{4}{c}{Sintel (train, clean)} & \multicolumn{4}{c}{Sintel (train, final)} & \multirow{2}{*}[-2pt]{\begin{tabular}[x]{@{}c@{}}Param\\(M) \end{tabular}}   & \multirow{2}{*}[-2pt]{\begin{tabular}[x]{@{}c@{}}Time\\(ms) \end{tabular}}  \\
    \addlinespace[-12pt] \\
    \cmidrule(lr){3-6} \cmidrule(lr){7-10} \cmidrule(lr){11-14}
    \addlinespace[-12pt] \\
    & & EPE & $s_{0-10}$ & $s_{10-40}$ & $s_{40+}$ & EPE & $s_{0-10}$ & $s_{10-40}$ & $s_{40+}$ & EPE & $s_{0-10}$ & $s_{10-40}$ & $s_{40+}$ & \\
    
    \midrule
    
    \multirow{6}{*}[-2pt]{RAFT \cite{teed2020raft}} & 0 & 14.28 & 1.47 & 3.62 & 40.48 & 4.04 & 0.77 & 4.30 & 26.66 & 5.45 & 0.99 & 6.30 & 35.19 & \multirow{6}{*}[-2pt]{5.3} & 25 (14) \\
    & 3 & 6.27 & 0.69 & 1.67 & 17.63 & 1.92 & 0.47 & 2.32 & 11.37 & 3.25 & 0.65 & 4.00 & 20.04 & & 39 (21) \\
    & 7 & 4.66 & 0.55 & 1.38 & 12.87 & 1.61 & 0.39 & 1.90 & 9.61 & 2.80 & 0.53 & 3.30 & 17.76 & & 58 (31) \\
    & 11 & 4.31 & 0.53 & 1.33 & 11.79 & 1.55 & 0.41 & 1.73 & 9.19 & 2.72 & 0.52 & 3.12 & 17.43 & & 78 (41) \\
    & 23 & 4.22 & 0.53 & 1.32 & 11.52 & 1.47 & 0.36 & 1.63 & 9.00 & 2.69 & 0.52 & 3.05 & 17.28 & & 133 (71)\\
    & 31 & 4.25 & \textbf{0.53} & 1.31 & 11.63 & 1.41 & 0.32 & 1.55 & 8.83 & 2.69 & 0.52 & 3.00 & 17.45 & & 170 (91) \\
    
    \midrule

     \multirow{2}{*}[-2pt]{{GMFlow}} & 0 & 3.48 & 0.67 & 1.31  & 8.97 & 1.50 & 0.46  & 1.77 & 8.26  & 2.96  & 0.72 & 3.45  & 17.70 & 4.7 & 57 (26) \\
     & 1 & \textbf{2.80} & \textbf{0.53} & \textbf{1.01} & \textbf{7.31} & \textbf{1.08} & \textbf{0.30} & \textbf{1.25} & \textbf{6.26} & \textbf{2.48} & \textbf{0.51}  & \textbf{2.81} & \textbf{15.67} & 4.7 & 151 (66) \\

    \bottomrule
    \end{tabular}
    \vspace{-5pt}
    \caption{\textbf{RAFT's iterative refinement framework \textit{vs.} our GMFlow framework}. The models are trained on Chairs and Things training sets. We use RAFT's officially released model for evaluation. The inference time is measured on a single V100 and A100 (in parentheses) GPU at Sintel resolution ($436\times 1024$). Our framework gains more speedup than RAFT ($2.29\times$ \vs. $1.87 \times$, \ie, ours: $151 \to 66$, RAFT: $170 \to 91$) on the high-end A100 GPU since our method doesn't require a large number of sequential computation.
    }
    \label{tab:raft_vs_ours}
    \vspace{-10pt}
\end{table*}

\subsection{Ablations}
\label{sec:ablation}

{\bf Transformer components.} We ablate different Transformer components in Table~\ref{tab:transformer}. The cross-attention contributes most, since it models the mutual relationship between two features, which is missing in the features extracted from the convolutional backbone network. Also, the position information makes the matching process position-dependent, and can be conducive to alleviate the ambiguities in pure feature similarity-based matching. Removing the feed-forward network (FFN) reduces a large number of parameters, while also leading to a moderate performance drop. The self-attention aggregates contextual cues within the same feature, bringing additional performance gains.

\begin{figure}[t]
    \centering
    \includegraphics[width=0.9\linewidth]{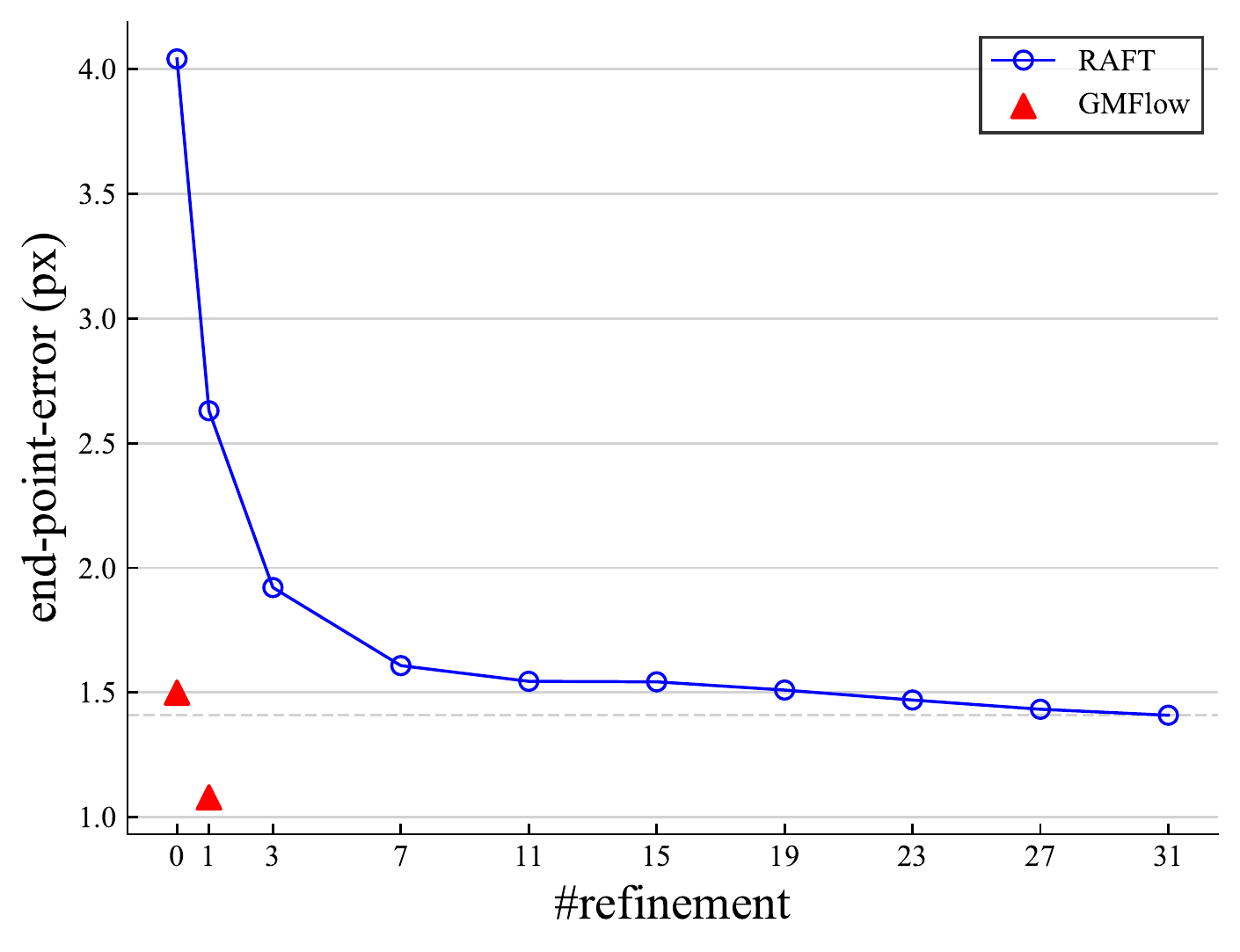}
    \vspace{-12pt}
    \caption{%
    Optical flow \textbf{end-point-error \vs number of refinements} at inference time. This figure shows the generalization results on Sintel (clean) training set after training on Chairs and Things datasets. 
    Our method outperforms 31-refinements RAFT's performance while using only one refinement and running faster.
    }
    \label{fig:iter_vs_epe}
    \vspace{-12pt}
\end{figure}

{\bf Local window attention.} We compare the speed-accuracy trade-off of splitting to different numbers of local windows for attention computation in Table~\ref{tab:split_attn}. Recall that the extracted features from backbone is $1/8$ resolution, further splitting to $H/2 \times W /2$ local windows (\ie, $1/16$ of the original resolution) represents a good trade-off between accuracy and speed, and thus is used in our framework.

{\bf Matching Space.} We replace our global matching with local matching in Table~\ref{tab:global_local_match} and observe a significant performance drop, especially for large motion ($s_{40+}$). Besides, the global matching can be computed efficiently with a simple matrix multiplication, while larger size for local matching will be slower due to the excessive sampling operation.

{\bf Flow propagation.} Our flow propagation strategy results in significant performance gains in unmatched regions (including occluded and out-of-boundary pixels), as shown in Table~\ref{tab:prop}. The structural correlation between the feature and flow provides a valuable clue to improve the performance of pixels that are challenging to match. Visual comparisons are provided in the \emph{supplementary material}.

{\bf Refinement.} We compare whether to share the backbone when extracting $1/8$ and $1/4$ features, and whether to share the Transformer for matching at $1/8$ and $1/4$ resolutions. The results are shown in Table~\ref{tab:refine}. Sharing Transformer and multi-scale features better regularizes the learning process, leading to better performance and less parameters.

{\bf Training length.} Our ablations thus far are based on 200K-iterations fine-tuning on Things dataset. In Table~\ref{tab:train_schedule}, we show our framework benefits from more training iterations, consistent with the observations in previous vision Transformer works \cite{dosovitskiy2020image,liu2021Swin}. We use 800K-iterations model as our final model in subsequent comparisons.

We analyze the computational complexities of core components in our framework in the \emph{supplementary material}.

\begin{table}[t]
    \centering
    \setlength{\tabcolsep}{1.5pt} %
    \begin{tabular}{llcccccccccccccc}
    \toprule
    
    Training data & Method & EPE & F1-all & $s_{0-10}$ & $s_{10-40}$ & $s_{40+}$ &   \\

    \addlinespace[-12pt] \\
    
    \midrule
    
    \multirow{2}{*}[-2pt]{C + T} & RAFT & 5.32 & 17.46 & 0.67 & 1.58 & 13.68 \\
    & GMFlow & 7.77 & 23.40 & 0.74 & 2.19 & 20.34 \\
    
    \midrule
    
    \multirow{2}{*}[-2pt]{C + T + VK} & RAFT & 2.45 & 7.90 & 0.43 & 1.18 & 5.70 \\
    & GMFlow & 2.85 & 10.77 & 0.49 & 1.16 & 6.87 \\

    \bottomrule
    \end{tabular}
    \vspace{-8pt}
    \caption{\textbf{Generalization on KITTI} after training on synthetic Chairs (C), Things (T) and Virtual KITTI 2 (VK) datasets. 
    }
    \label{tab:gen_kitti}
    \vspace{-16pt}

\end{table}

\begin{figure*}[t]
    \centering
    \includegraphics[width=0.98\linewidth]{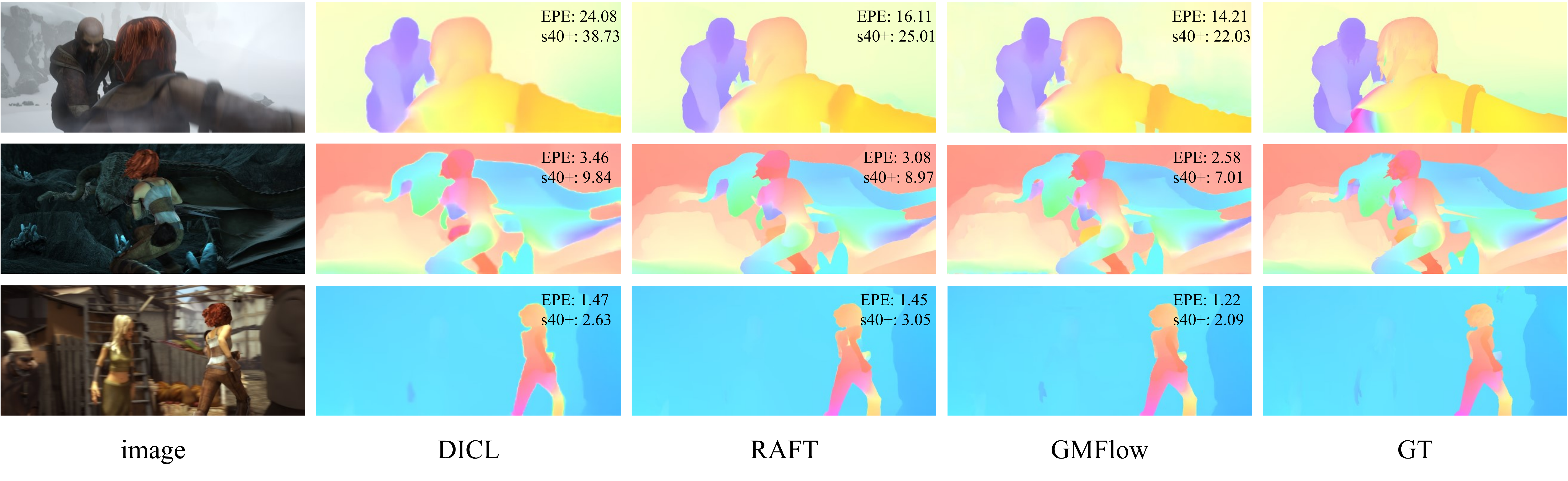}
    \vspace{-8pt}
    \caption{\textbf{Visual comparisons on Sintel test set}.}
    \label{fig:sintel_test}
    \vspace{-12pt}
\end{figure*}

\subsection{Comparison with RAFT}
\label{sec:compare_raft}

{\bf Sintel.} Table~\ref{tab:raft_vs_ours} shows the results on Things validation set and Sintel clean and final training sets after training on Chairs and Things training sets. Without using any refinement, our method achieves better performance on Things and Sintel (clean) than RAFT with 11 refinements. By using an additional refinement, our method outperforms RAFT with 31 refinements, especially on large motion ($s_{40+}$). Fig.~\ref{fig:iter_vs_epe} visualizes the results. Furthermore, our framework enjoys faster inference speed compared to RAFT and also does not require a large number of sequential processing. On the high-end A100 GPU, our framework gains more speedup compared with RAFT's sequential framework ($2.29\times$ \vs. $1.87 \times$, \ie, ours: $151 \to 66$, RAFT: $170 \to 91$), reflecting that our framework can benefit more from advanced hardware acceleration and showing its potential for further speed optimization.

{\bf KITTI.} Table~\ref{tab:gen_kitti} shows the generalization results on KITTI training set after training on Chairs and Things training sets. In this evaluation setting, our framework doesn't outperform RAFT, which is mainly caused by the gap between the synthetic training sets and the real-world testing dataset. One key reason behind our inferior performance is that RAFT, relying on fully convolutional neural networks, benefits from the inductive biases in convolution layers, which requires a relatively smaller size training data to generalize to a new dataset in comparison with Transformers \cite{dosovitskiy2020image, d2021convit,xu2021vitae,zhang2022vitaev2}. To substantiate this claim, we fine-tune both RAFT and our method on the additional Virtual KITTI 2 \cite{cabon2020virtual} dataset. We can see from Table~\ref{tab:gen_kitti} that the performance gap becomes smaller when more data is available.

\begin{table}[t]
\footnotesize
    \centering
    \setlength{\tabcolsep}{2pt} %
    \begin{tabular}{lcccccc}
    \toprule
    \multirow{2}{*}[-2pt]{Method } & \multicolumn{3}{c}{Sintel (clean)} &
    \multicolumn{3}{c}{Sintel (final)} \\
    \addlinespace[-12pt]  \\
    \cmidrule(lr){2-4} \cmidrule(lr){5-7} 
    \addlinespace[-12pt] \\
    & all & matched & unmatched & all & matched & unmatched \\
    \midrule
    
    FlowNet2 \cite{ilg2017flownet} & 4.16 & 1.56 & 25.40 & 5.74 & 2.75 & 30.11 \\
    PWC-Net+ \cite{sun2019models} & 3.45 & 1.41 & 20.12 & 4.60 & 2.25 & 23.70  \\
    HD$^3$ \cite{yin2019hierarchical} & 4.79 & 1.62 & 30.63 & 4.67 & 2.17 & 24.99 \\
    VCN \cite{yang2019volumetric} & 2.81 & 1.11 & 16.68 & 4.40 & 2.22 & 22.24 \\
    DICL \cite{wang2020displacement} & 2.63 & 0.97 & 16.24 & 3.60 & 1.66 & 19.44 \\
    
    RAFT \cite{teed2020raft} & 1.94 & - & - & 3.18 & - & - \\
    GMFlow & \textbf{1.74} & \textbf{0.65} & \textbf{10.56} & \textbf{2.90} & \textbf{1.32} & \textbf{15.80} \\
    
    \midrule
    
    RAFT$^\dagger$ \cite{teed2020raft} & 1.61 & 0.62 & 9.65 & 2.86 & 1.41 & 14.68 \\
    GMA$^\dagger$ \cite{Jiang_2021_ICCV} & 1.39 & 0.58 & 7.96 & 2.47 & 1.24 & 12.50 \\

    \bottomrule
    \end{tabular}
    \vspace{-8pt}
    \caption{\textbf{Comparisons on Sintel test test.} $^\dagger$ represents the method uses last frame's flow prediction as initialization for subsequent refinement, while other methods all use two frames only.}
    \vspace{-16pt}
    \label{tab:sintel_compare}
\end{table}

\subsection{Comparison on Benchmarks}
\label{sec:sintel_kitti}

{\bf Sintel.} Following previous works \cite{teed2020raft,xu2021high}, we further fine-tune our Things model on several mixed datasets that consist of KITTI \cite{menze2015object}, HD1K \cite{kondermann2016hci}, FlyingThings3D \cite{mayer2016large} and Sintel \cite{butler2012naturalistic} training sets. We perform fine-tuning for 200K iterations with a batch size of 8 and a learning rate of 2e-4. The results on Sintel test set are shown in Table~\ref{tab:sintel_compare}. We achieve best performance among all the competitive state-of-the-art methods that only use two frames at inference. Although RAFT can also use last frame's prediction as initialization for subsequent refinement, we still achieve comparable performance in matched regions even without using multi-frame information. Qualitative comparisons between GMFlow and other state-of-the-art approaches on Sintel test set are shown in Fig.~\ref{fig:sintel_test}. More visual results on DAVIS \cite{perazzi2016benchmark} dataset are given in the \emph{supplementary material}.

{\bf KITTI.} We perform additional fine-tuning on the KITTI 2015 training set from the model trained on Sintel. We train GMFlow for 100K iterations with a batch size of 8 and a learning rate of 2e-4. Table~\ref{tab:kitti_test_set} shows the evaluation results. For non-occluded (Noc) pixels, our performance is slightly inferior to RAFT. For all pixels that contain both non-occluded and occluded pixels, the performance gap becomes larger, which indicates that our inferior performance largely lies in occluded regions.

\begin{table}[t]
\footnotesize
    \centering
    \setlength{\tabcolsep}{3.pt} %
    
    \begin{tabular}{lccccccccccccc}
    \toprule

    Method & FlowNet2 \cite{ilg2017flownet} & PWC-Net+ \cite{sun2019models} & RAFT \cite{teed2020raft} & GMFlow \\
    
    \midrule
    
    All & 11.48 & 7.72 &  5.10 & 9.32 \\
    Noc & 6.94 & 4.91 & 3.07 & 3.80 \\

    \bottomrule
    \end{tabular}
    \vspace{-8pt}
    \caption{\textbf{Comparisons on KITTI test set}. The metric is F1-all. ``All'' denotes the evaluation results on all pixels with ground truth, and ``Noc" denotes non-occluded pixels only.}
    \label{tab:kitti_test_set}
    \vspace{-16pt}
\end{table}

\subsection{Limitation and Discussion}

Our framework still has room for future improvement in occluded regions, as can be seen from the KITTI results in Table~\ref{tab:kitti_test_set}.
Besides, our framework may not generalize very well when the training data has significantly large gap with the test data (\eg, synthetic Things to real-world KITTI). 
Fortunately, there are many large-scale datasets available currently, \eg, Virtual KITTI \cite{gaidon2016virtual,cabon2020virtual}, VIPER \cite{richter2017playing}, REFRESH \cite{lv2018learning}, AutoFlow \cite{sun2021autoflow} and TartanAir \cite{tartanair2020iros}, they can be used to enhance Transformer's generalization ability.

\section{Conclusion}

We have presented a new global matching formulation for optical flow and demonstrated its strong performance.
We hope our new perspective will pave a way towards a new paradigm for accurate and efficient optical flow estimation.

\textbf{Broader impact.} Our proposed method might produce unreliable results in occluded regions, thus care should be taken when using the prediction results from our model, especially for safety-critical scenarios like self-driving cars.

{\bf Acknowledgement.} This research is supported in part by Monash FIT Start-up Grant. Dr. Jing Zhang is supported by ARC FL-170100117.

{\small
\bibliographystyle{ieee_fullname}
\bibliography{egbib}
}

\newpage

\section*{Appendix}
\renewcommand{\thesection}{\Alph{section}}
\renewcommand{\thetable}{\Alph{table}}
\renewcommand{\thefigure}{\Alph{figure}}
\setcounter{section}{0}
\setcounter{table}{0}
\setcounter{figure}{0}

\section{More Comparisons}

We present more comprehensive comparisons (as a supplement of Table~\ref{tab:conv_vs_softmax} in the main paper) with all the possible combinations of flow estimation approaches in Table~\ref{tab:conv_vs_softmax_add}. Our Transformer and softmax-based method is consistently better and has less parameters than other variants.

\begin{table*}[t]
    \centering
    \setlength{\tabcolsep}{3pt} %
    \begin{tabular}{lcccccccccccccccccc}
    \toprule
    \multirow{2}{*}[-2pt]{Method} & \multirow{2}{*}[-2pt]{\begin{tabular}[x]{@{}c@{}}feature\\enhancement \end{tabular}} & \multirow{2}{*}[-2pt]{\begin{tabular}[x]{@{}c@{}}flow \\ prediction \end{tabular}} & \multirow{2}{*}[-2pt]{\#convs} & \multicolumn{3}{c}{Sintel (train, clean)} & \multicolumn{3}{c}{Sintel (train, final)} &  \multirow{2}{*}[-2pt]{\begin{tabular}[x]{@{}c@{}}Param\\(M) \end{tabular}} \\
    \cmidrule(lr){5-7} \cmidrule(lr){8-10} 
    & & & & all & matched & unmatched & all & matched & unmatched & \\
    
    \midrule
    
    \multirow{4}{*}[-2pt]{variants} & - & cost + conv & 14 & 3.32 & 1.56 & 22.34 & 4.93 & 2.82 & 27.73 & 4.64 \\
    & Transformer & cost + conv & 2 & 3.41 & 2.40 & 14.32 & 4.57 & 3.27 & 18.67 & 4.95 \\
    & Transformer & cost + conv & 14 & 2.04 & 1.09 & 12.34 & 3.37 & 2.03 & 17.80 & 7.79 \\
    & conv & softmax & 14 & 6.36 & 3.22 & 40.30 & 8.00 & 4.80 & 42.58 & 5.12 \\

    \midrule
    
    GMFlow (w/o prop.) & Transformer & softmax & 0 & 2.28 & \textbf{1.06} & 15.54 & 3.44 & \textbf{1.95} & 19.50 & 4.20 \\ 
    GMFlow (w/ prop.) & Transformer & softmax & 0 & \textbf{1.89} & 1.10 & \textbf{10.39} & \textbf{3.13} & 1.98 & \textbf{15.52} & 4.23 \\
    
    \bottomrule
    \end{tabular}
    \caption{\textbf{Comparisons on different variants of flow estimation approaches.} Although the Transformer can also be used for feature enhancement in the cost volume and convolution-based approach (cost + conv), its performance heavily relies on a deep convolutional regressor (\eg, 14 layers to catch up). In contrast, our softmax-based method is \emph{parameter-free} (4.20M \vs. 7.79M). The flow propagation (prop.) layer further improves ours performance in unmatched regions, while only introducing additional 0.03M parameters. Replacing the Transformer with convolutions for feature enhancement leads to significantly large performance drop, since convolutions are not able to model the mutual relationship between two features.
    }
    \label{tab:conv_vs_softmax_add}
    \vspace{-6pt}
\end{table*}

\section{Computational Complexity}

We analyze the computational complexities of core components in our framework below.

{\bf Global Matching.} In our global matching formulation, we build a 4D correlation matrix $H \times W \times H \times W$ to model all pair-wise similarities between two features (with size $H \times W$, $1/8$ of the original image resolution). There exists an equivalent implementation should it become a bottleneck for high-resolution images. Note that the pixels in the first feature are \emph{independent} and thus their flow predictions can be computed \emph{sequentially}. Specifically, we can sequentially compute $K \times K$ correlation matrices (each with size $H/K \times W/K \times H \times W$), and finally merge the results for all pixels. Such a sequential implementation can save the memory consumption while having little influence on the overall inference time (see Table~\ref{tab:seq_global_match}), since the global matching operation only needs to compute once, and it's not a significant speed bottleneck in the full framework.

\begin{table}[H]
    \centering
    \setlength{\tabcolsep}{3.pt} %
    \begin{tabular}{cccccccccccccccc}
    \toprule
    
    \#splits & $1 \times 1$ & $2 \times 2$ & $4 \times 4$ & $8 \times 8$ \\
    
    \midrule
    
    Time (ms) & 52.57 & 52.64 & 52.90 & 59.45 \\
    
    \bottomrule
    \end{tabular}
    \vspace{-5pt}
    \caption{\textbf{Inference time \vs. number of splits for sequential global matching implementation.} The input image resolution is $448 \times 1024$, and the features are downsampled by $8 \times$.
    }
    \label{tab:seq_global_match}
    \vspace{-10pt}
    
\end{table}

{\bf Transformer.} We use shifted local window attention \cite{liu2021Swin} in the Transformer implementation, where each local window size is $1/16$ of the original image resolution by default. The computational cost is usually acceptable for regular image resolutions (\eg, $448 \times 1024$). Note that we can always switch to smaller windows size (\eg, $1/32$, see Table~\ref{tab:split_attn} of the main paper) should it become a bottleneck.

{\bf Flow Propagation.} Our default flow propagation scheme computes a global self-attention. The sequential implementation in global matching can also be adopted here. It's also possible to compute a local window self-attention only for less memory consumption by trading some accuracy in large motion (Table~\ref{tab:global_local_prop}). Such a local attention operation can be implemented efficiently with PyTorch's \texttt{unfold} function.

\begin{table}[H]
    \centering
    \setlength{\tabcolsep}{3.pt} %
    \begin{tabular}{cccccccccccccccc}
    \toprule
    
    \multirow{2}{*}[-2pt]{self-attn.} & \multicolumn{4}{c}{Sintel (train, final)}  & \\
    \addlinespace[-12pt] \\
    \cmidrule(lr){2-5} 
    \addlinespace[-12pt] \\
    & EPE & $s_{0-10}$ & $s_{10-40}$ & $s_{40+}$  \\
    \midrule
    
    global & \textbf{3.13} & 0.80 & 3.87 & \textbf{18.04} \\
    local $3 \times 3$ & 3.31 & 0.79 & 3.75 & 20.22 \\
    local $5 \times 5$ & 3.21 & \textbf{0.75} & \textbf{3.66} & 19.69 \\

    \bottomrule
    \end{tabular}
    \vspace{-5pt}
    \caption{\textbf{Global \vs. local self-attention for flow propagation.}
    }
    \label{tab:global_local_prop}
    \vspace{-10pt}
    
\end{table}

{\bf Refinement.} Although the feature resolution of our refinement architecture is higher ($1/4$), it is not a significant bottleneck since smaller local window ($1/32$ of the original image resolution) attention is used in the Transformer and matching is performed within a local window.

Overall, our GMFlow framework is general and flexible, and many concrete implementations are possible to meet specific needs.

\section{More Visual Results}

\begin{figure*}[t]
    \centering
    \includegraphics[width=0.98\linewidth]{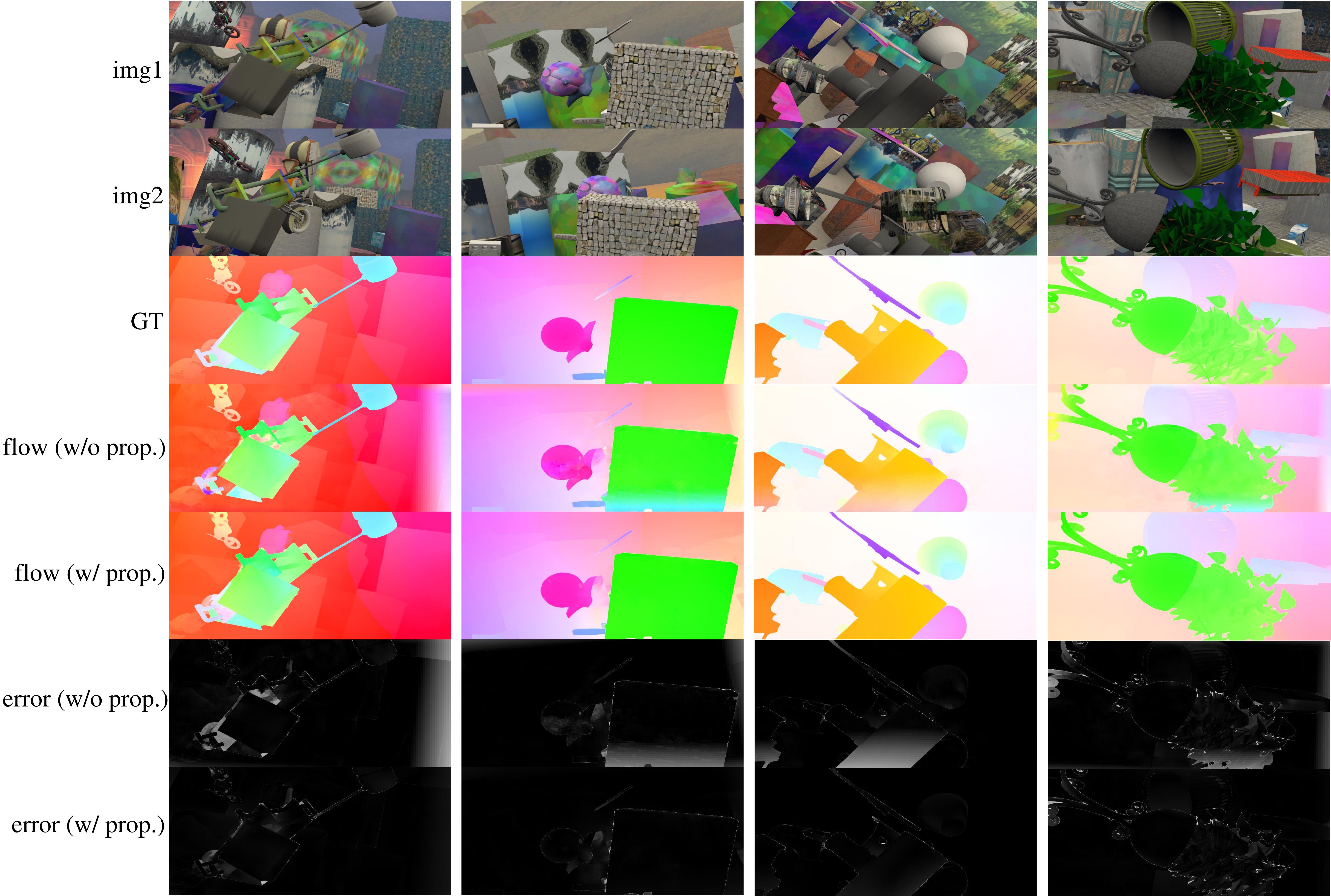}
    \caption{Our \textbf{flow propagation} (prop.) scheme significantly improves the performance of occluded and out-of-boundary pixels.}
    \label{fig:flow_prop}
\end{figure*}

{\bf Flow Propagation.} Our flow propagation scheme with self-attention is quite effective for handling occluded and out-of-boundary pixels, as can be seen from Fig.~\ref{fig:flow_prop}.

\begin{figure*}[t]
    \centering
    \includegraphics[width=0.98\linewidth]{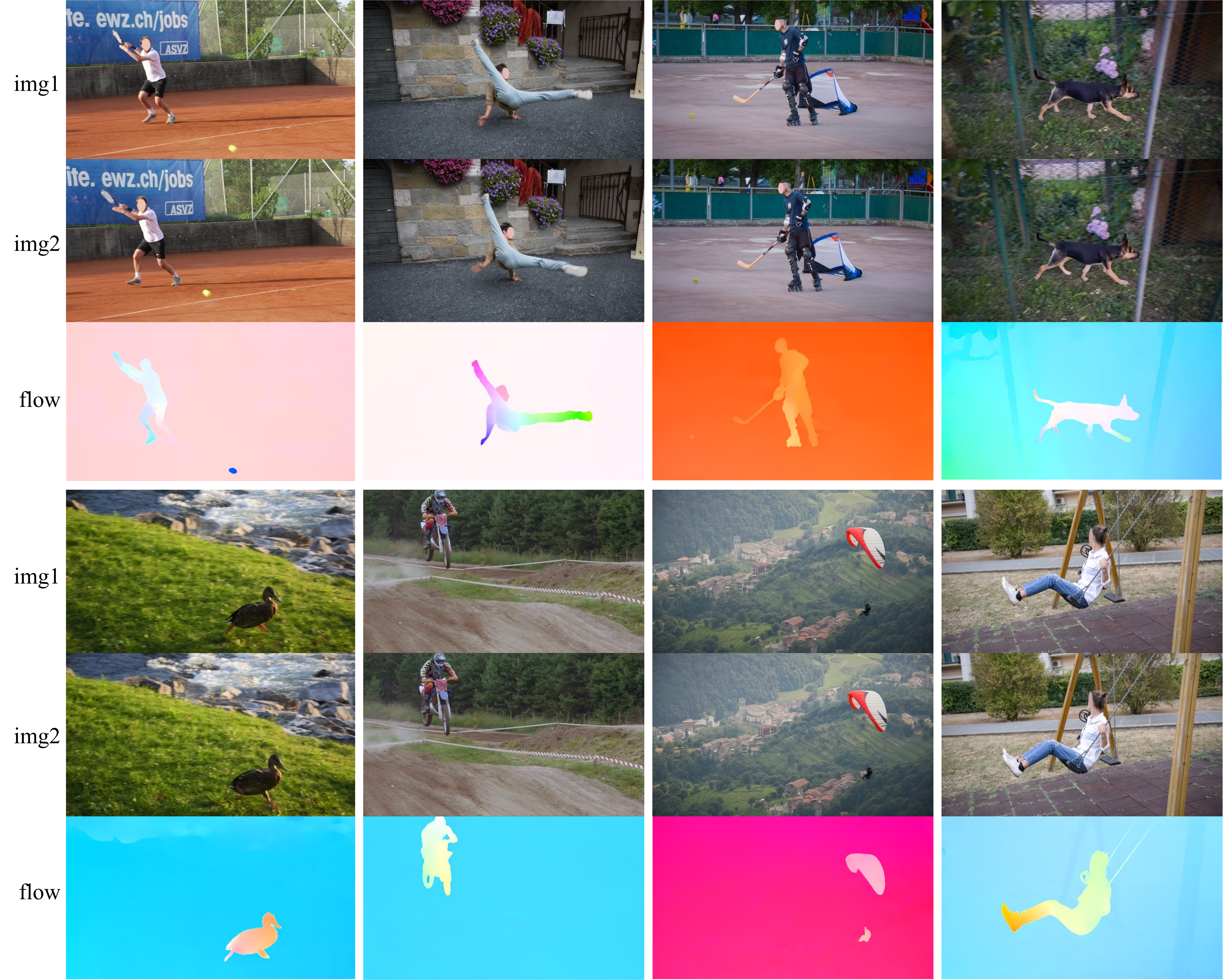}
    \caption{\textbf{Visual results on DAVIS dataset}.}
    \label{fig:vis_davis}
\end{figure*}

{\bf Prediction on DAVIS dataset.} We test our pre-trained Sintel model on the DAVIS \cite{perazzi2016benchmark} dataset, the results on diverse scenes are shown in Fig.~\ref{fig:vis_davis}.

\section{More Implementation Details}

{\bf Network Architectures.} The Transformer feature dimension is 128, and the intermediate feed-forward network expands the dimension by $4\times$. We only use a single head in all the attention computations, since we observe that multi-head attention slows down the speed without bringing obvious performance gains. Our refinement architecture uses exactly the same Transformer for feature enhancement, except that the attentions are performed within smaller local windows. The self-attention layer in the flow propagation step is also shared for $1/8$ and $1/4$ resolutions, where we perform global attention at $1/8$ resolution and local $3 \times 3$ window attention at $1/4$ resolution.

{\bf Training Details.} Our data augmentation strategy mostly follows RAFT \cite{teed2020raft} except that we didn't use occlusion augmentation, since no obvious improvement is observed in our experiments. During training, we perform random cropping following previous works. The crop size for FlyingChairs is $384 \times 512$, FlyingThings3D is $384 \times 768$, Sintel is $320 \times 896$ and KITTI is $320 \times 1152$. Our framework without refinement is trained on 4 V100 (16GB) GPUs. The full framework with refinement is trained on 4 A100 (40GB) GPUs. We are also able to reproduce the results on 4 V100 (16GB) GPUs by halving the batch size and doubling the training iterations.

\end{document}